\documentclass[sigconf]{acmart}

\usepackage{xcolor}
\usepackage{subcaption}
\usepackage{booktabs}
\usepackage{multirow}
\usepackage{amsmath}

\copyrightyear{2021}
\acmYear{2021}
\setcopyright{rightsretained}
\acmConference[AIES '21]{Proceedings of the 2021 AAAI/ACM Conference on AI, Ethics, and Society}{May 19--21, 2021}{Virtual Event, USA}
\acmBooktitle{Proceedings of the 2021 AAAI/ACM Conference on AI, Ethics, and Society (AIES '21), May 19--21, 2021, Virtual Event, USA}\acmDOI{10.1145/3461702.3462594}
\acmISBN{978-1-4503-8473-5/21/05}

\newcommand{\person}{\emph{person}}
\newcommand{\man}{\emph{man}}
\newcommand{\woman}{\emph{woman}}
\newcommand{\boy}{\emph{boy}}
\newcommand{\girl}{\emph{girl}}

\begin{document}
\fancyhead{}

\title{A Step Toward More Inclusive People Annotations for Fairness}

\author{Candice Schumann}
\affiliation{%
  \institution{Google}
  \country{United States}}
\email{cschumann@google.com}

\author{Susanna Ricco}
\affiliation{%
  \institution{Google}
  \country{United States}}
\email{ricco@google.com}

\author{Utsav Prabhu}
\affiliation{%
 \institution{Google}
 \country{United States}}
\email{utsavprabhu@google.com}

\author{Vittorio Ferrari}
\affiliation{%
  \institution{Google}
  \country{Switzerland}}
\email{vittoferrari@google.com}

\author{Caroline Pantofaru}
\affiliation{%
  \institution{Google}
  \country{United States}}
\email{cpantofaru@google.com}

\begin{abstract}

The Open Images Dataset \cite{Kuznetsova20:Open} contains approximately 9 million images and is a widely accepted dataset for computer vision research. As is common practice for large datasets, the annotations are not exhaustive, with bounding boxes and attribute labels for only a subset of the classes in each image. In this paper, we present a new set of annotations on a subset of the Open Images dataset called the \emph{MIAP (More Inclusive Annotations for People)} subset, containing bounding boxes and attributes for all of the people visible in those images. The attributes and labeling methodology for the \emph{MIAP} subset were designed to enable research into model fairness. In addition, we analyze the original annotation methodology for the \person{} class and its subclasses, discussing the resulting patterns in order to inform future annotation efforts. By considering both the original and exhaustive annotation sets, researchers can also now study how systematic patterns in training annotations affect modeling.

\end{abstract}




\maketitle

\section{Introduction}
The Open Images Dataset \cite{Kuznetsova20:Open} contains approximately 9 million images with annotations supporting research into image classification, object detection and instance segmentation, visual relationship detection, and spatially localized voice and text descriptions of the image contents. It is widely accepted as a key dataset to both measure and advance the state of the art in image and scene understanding. 

In this paper, we introduce a new set of annotations, released as the \emph{MIAP (More Inclusive Annotations for People)} collection in Open Images Extended, which improve the bounding box annotations for the five original classes in the \person{} subtree. The annotations use a new labeling protocol designed to produce annotations that enable fairness analysis. With the increasing focus on reducing unfair bias as part of responsible AI research, we hope these annotations will encourage researchers already leveraging Open Images to incorporate fairness analysis in their research.

Open Images contains image-level labels for more than 19,000 classes, and bounding box annotations for a subset of 600 of those classes.
Even for those 600 classes, it is infeasible to exhaustively label each of them as present (positive) or absent (negative) in every image.
Hence, Open Images annotations are not exhaustive, as is common practice for large scale computer vision datasets, e.g.~\cite{Lin14:Microsoft,Deng09:Imagenet}. Each image is associated with a small set of positive labels and a small set of negative labels. These labels are proposed by an ensemble of image classifiers (e.g., labels produced with confidence above a threshold by some algorithm) and then verified as either positive or negative by a human annotator. The presence/absence of any label not proposed by the initial image classifier is simply unknown.

After this stage of producing image-level labels, annotators then draw bounding boxes for the classes at the most specific level of the 600-class hierarchy (leaves) that are included in the image-level positive label set.

Of the 600 classes with bounding-boxes, five classes are part of the \person{} hierarchy: \person, \man, \woman, \boy, and \girl.

Here, Open Images departs from other large-scale object detection datasets~\cite{Lin14:Microsoft,ILSVRC15} by including these subclasses.
Although most applications require annotations for only the encompassing \person{} class, the availability of gender- and age range-specific labels would in theory allow for use of these labels for fairness analysis and bias mitigation. However, we found the specific labeling procedure used impacted the accuracy of these subgroup labels and resulted in non-exhaustive bounding box annotations for the \person{} superclass for some images.

Our new \emph{MIAP} annotations collapse the gender- and age range-specific subclasses into a single \person{} superclass at annotation time.
We ask annotators to draw bounding boxes around \emph{all} people appearing in an image with any positive image-level label of either \person, \man, \woman, \boy, or \girl.
\emph{Person} boxes are then annotated with additional attributes corresponding to the perceived age range and perceived gender presentation of the individual. This procedure adds a significant number of boxes that were previously missing due to complex interactions between the scene content, the characteristics of the individual as interpreted by the initial image classifier, and the perception of two sets of human annotators.
Our boxes give more complete ground truth for training a person detector as well as more accurate subgroup labels for fairness analysis and bias mitigation.

\subsection{Summary of Contributions}
The main contributions of this paper are:
\begin{itemize}
    \item An analysis of the original Open Images annotation pipeline's impact on the completeness of person-related box annotations.
    \item An updated annotation procedure to produce more complete annotations for the \person{} class.
    \item The resulting annotations on a subset of images and a comparative analysis of the differences between the two annotation types. We publicly release our new annotations.
\end{itemize}

\subsection{Intended Use}
We include annotations for perceived age range and gender presentation for \person{} bounding boxes because we believe these annotations are necessary to advance our field's ability to better understand and work to mitigate and eliminate unfair bias or disparate performance across protected subgroups. We note that the labels capture the gender and age range presentation as assessed by a third party based on visual cues alone, rather than an individual's self-identified gender or actual age. We do not support or condone building or deploying gender and/or age presentation classifiers trained from these annotations as we believe the risks associated with the use of these technologies outside fairness research outweigh any potential benefits. We refer readers to the excellent work detailing these potential risks (\emph{e.g.}, \cite{Hamidi2018:Gender,Keyes2018:Misgendering}) for more information.

\section{Related Work}
Research into techniques to improve fairness in computer vision algorithms is of increasing importance, as these systems become widely used. Existing approaches span the model development cycle, including (a) careful formulation of the task, (b) identifying and reducing potential biases in training datasets, (c) inventing techniques for training fairer models in spite of biased datasets, (d) defining and evaluating whether the resulting system achieves some desired fairness property. This work is mainly focused on (b) and (d).

\subsection{Identifying and reducing bias in datasets}
Bias can manifest in a dataset as unintended consequence of design choices made during dataset construction. Datasets containing people are particularly sensitive to these biases, and can have harmful downstream implications - such as resulting in widely-used systems that have unequal performance across population demographics. For example, datasets that sample images from internet sources may over-represent some geographical regions compared to others. Both Open Images and its predecessor ImageNet have been shown to contain a disproportionate number of images from North America and Europe~\cite{Shankar2017,Vries_2019_CVPR_Workshops}. Classifiers trained on this distribution were more likely to misclassify images from the developing world for both person-related classes (e.g., \emph{bridegroom})~\cite{Shankar2017} and object classes (e.g., \emph{spices})~\cite{Vries_2019_CVPR_Workshops}.

To partially address this limitation and encourage further research~\cite{Atwood2018:Inclusive}, Open Images now includes the Crowdsourced Extension\footnote{https://storage.googleapis.com/openimages/web/extended.html}, containing new images from countries underrepresented in the original set. This Open Images extension is focused on image-level labels for non-human classes. The faces of people appearing in this set are blurred, making it less suitable for analysis of person detectors. 

Of course, the potential for bias in large web-scale datasets goes far beyond skewed geographic distributions. The exceptional work updating the person subtree of ImageNet~\cite{Yang2020:ImageNet} provides detailed analysis of sources of bias, from the definitions of class labels to the distribution of people representing given classes. Their work focuses on correcting image-level labels while we chose to focus on the bounding box annotations for people in Open Images. With our updated annotations, we provide the attribute labels similar to those annotated at the image level in ImageNet. We hope future research can leverage both datasets to build on this work, improving these and future large-scale computer vision datasets.

\subsection{Identifying and reducing bias in trained models}
Creating better balanced and less biased datasets are steps toward the eventual goal of ML models that work well for everyone, whose predictions do not unfairly disadvantage any segment of the population. Most definitions of model fairness (\emph{e.g.}, equality of opportunity \cite{Hardt2016}) require partitioning an evaluation set by the attributes to which model performance should be agnostic. Proposed standards for model transparency reporting \cite{Zaldivar2019:Model}, which complement proposed transparency reports for datasets~\cite{gebru2018datasheets}, advocate for always reporting performance metrics broken down by relevant population groups, independent of specific fairness goals. Finally, many bias mitigation strategies require attribute labels on at least a portion of the model's training set~\cite{beutel2019putting,madras2018learning}.

There is a rich history of datasets containing images of people that include attribute annotations such as perceived gender and/or age. For some, these annotations were added for the explicit purpose of furthering ML fairness research. For example, the Pilot Parliaments Benchmark from Gender Shades~\cite{buolamwini2018gender} used gender and skin type annotations to highlight disparate performance of available gender classifiers across intersectional groups. Similarly, datasets for training or evaluating face recognition algorithms may include annotations for fairness analysis~\cite{watson2016nist,grother2018ongoing}. In other cases, these annotations were initially curated as training data for attribute classifiers (e.g., Celeb-A \cite{liu2015faceattributes}, FairFace \cite{karkkainenfairface}, UTKFace \cite{zhifei2017cvpr}) but may also be useful for studying bias~\cite{ryu2017inclusivefacenet,Alvi18}.

The attribute labels we provide differ from previous datasets in two key areas. First, our annotations do not rely on machine learned models for annotating either perceived gender or age range. Other large-scale annotation efforts exploring bias in ImageNet representation~\cite{Dulhanty2019:Auditing} or attempting to quantify the diversity in appearance across the population~\cite{merler2019diversity} resort to attribute labels produced by trained attribute classifiers. We prefer to reduce the influence of potentially biased attribute classifiers, especially when using these annotations for fairness applications. Second, and importantly, most other datasets with similar annotations for fairness applications are targeted toward either face attribute estimation or face recognition as their primary applications. As a result, they contain a single face of interest per image. In contrast, we augment a portion of the existing Open Images dataset, thus adding fairness annotations to images with multiple people per image, including for people whose faces are not clearly visible.

To our knowledge, the only other dataset containing multiple people in complex scenes that has been augmented with attribute annotations for the purpose of fairness research is BDD100k~\cite{Wilson2019:Predictive}. This work annotates Fitzpatrick skin type for localized pedestrians of sufficient size in the images. Skin tone is an important attribute to consider for fairness, and one that our work does not address. While both datasets can be used for evaluating whether a person detector performs consistently across population slices, BDD100k is limited to self-driving applications. In contrast, our annotations also provide data for fairness analysis of a broad set of scene understanding tasks.

\begin{figure*}
\centering
\includegraphics[width=0.9\textwidth]{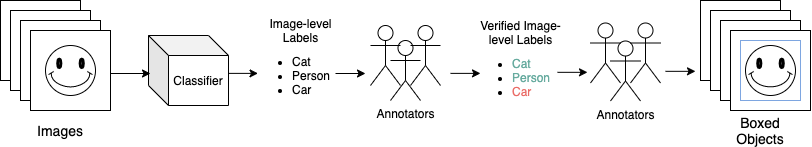}
\caption{The Open Images annotation pipeline from raw images to labeled object bounding boxes.}
\label{fig:pipeline}
\end{figure*}

\section{Analysis of the Original Open Images}
In this section we take a deep dive into the effects of the partially annotated ground truth label generation process on the annotation of people boxes.
Prior to this deep dive we do want to mention that the original Open Images dataset functions correctly as a standard partial ground truth dataset.
However, due to the non-exhaustive nature of the annotations, fairness evaluations or bias mitigation using this dataset could provide potentially misleading results.

\subsection{The Annotation Pipeline}
\label{sec:pipeline}

Open Images employs a two-stage process, first annotating image-level labels and then object bounding-boxes based on them (Figure~\ref{fig:pipeline}). In total, while Open Images considers a very large set of 19,794 possible classes at the image-level, object bounding-boxes are annotated for 600 of them. We focus on these from now on.

Given an image, an image classifier is used to generate potential positive and negative labels. These potential labels are then sent to annotators to verify. Each potential label is either marked as positive (i.e. the named object class is present in the image) or negative (i.e. the named object class does not appear in the image).
The presence or absence of any label that is not proposed by the image classifier is considered simply {\em unknown} (i.e. it is neither positive nor negative).

\begin{figure}
    \centering
    \includegraphics[width=0.6\columnwidth]{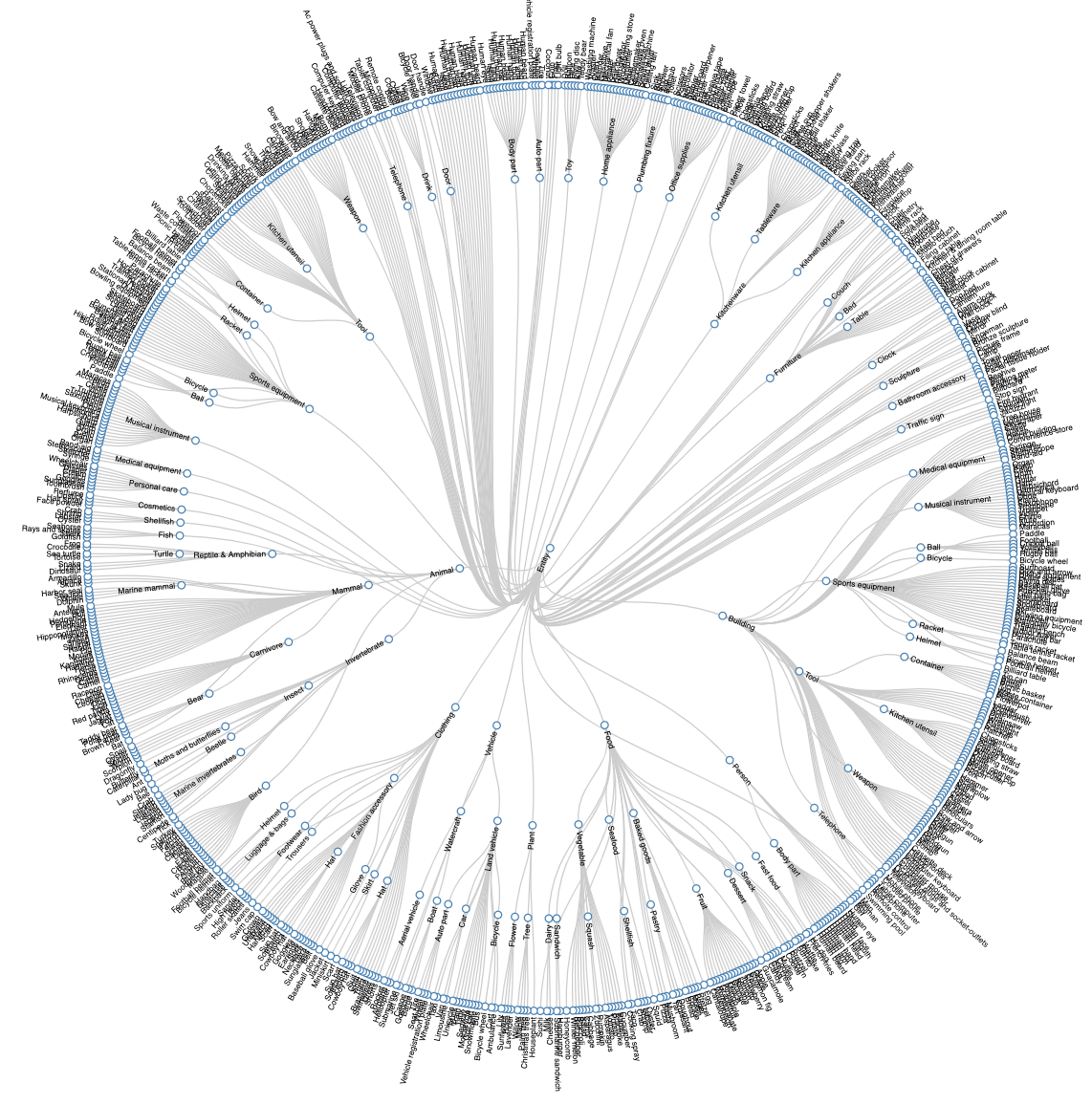}
    \caption{Visualization of the Open Images hierarchy of classes that are boxed~\cite{Kuznetsova20:Open} to demonstrate its scope and complexity.}
    \label{fig:full_hierarchy}
\end{figure}

Annotators are then asked to draw bounding-boxes around objects for the positive image labels, focusing on the most specific labels in the hierarchy (Figure~\ref{fig:full_hierarchy}). For example, if an image has positive labels of ``car", ``limousine", and ``screwdriver", the annotators box all limousines and screwdrivers (since limousine is more specific than car in the hierarchy). This means that if the image contains a car that is not a limousine, there will not be a box drawn around it.
Moreover, there will also be no boxes for classes that are unknown to be present or absent in the image (i.e. not proposed by the initial classifier, see above).

Note that this partial annotation protocol is working as intended -- it would be a huge undertaking to box everything in every image and partial labeling is standard practice~\cite{Kuznetsova20:Open,Deng09:Imagenet,Lin14:Microsoft}.
The Open Images protocol enables users to reconstruct the list of classes for which boxes might be missing, and take this into account when training or evaluating detectors (i.e. all unknown labels for an image, plus all positive labels which are not the most specific for that image).
That being said, partial annotation can have unintended consequences, especially when annotating people.

\subsection{Annotation Pipeline Mixing with Societal Norms}\label{sec:societal_norms}
Many standard academic datasets for object detection (such as MSCOCO~\cite{Lin14:Microsoft} and ImageNet~\cite{ILSVRC15}) contain a single \person{} class. Instead, the Open Images hierarchy contains five classes that correspond to localized people: \person, \man, \woman, \boy, and \girl.
While these gender- and age-specific labels are relevant and useful for some applications (such as fairness evaluations or bias mitigation), for others, the additional specificity is unnecessary and even undesirable.

The most obvious approach for training a detector with a single gender- and age-agnostic \person{} class would be to simply drop \man, \woman, \boy, \girl{} from the label set. However, because the hierarchical labeling process preferred labeling for a subclass when both superclass and subclass appeared in the positive image-level labels, dropping annotations for these four subclasses will result in dropping many ground truth boxes. For example, a training image containing a single man, with positive image-level labels for both \person{} and \man{} will contain a box labeled as \man{} but no corresponding box labeled as \person{} will exist. 

Another option is to collapse the hierarchy to a single class, treating boxes labeled as \man, \woman, \boy, \girl{} as \person{} boxes. This remapping could happen as a preprocessing step before training or as postprocessing at inference time. For the moment, we focus on remapping prior to training and defer a discussion of postprocessing to Section~\ref{sec:impact}.

The previous example is resolved correctly with label preprocessing, with the \man{} box remapped to \person{} providing correct ground truth for that image. 
Unfortunately, this procedure does not always result in a complete set of bounding boxes at training time, which requires localizing every person in the image. 
For a person appearing in an image to have a corresponding bounding box after label remapping given the original annotation process, one of two things had to be true:
\begin{itemize}
    \item The image classifier proposed \person{} as a potential image-level label for the image without proposing any gender- and age-specific subclass that the annotators marked as positive, \emph{or}
    \item{The image classifier must have proposed the subclass that both the image-level annotator and box-level annotators agreed described the specific person.}
\end{itemize}
We show a few examples below where these conditions are not satisfied.

\begin{figure}
    \centering
    \includegraphics[width=0.7\columnwidth]{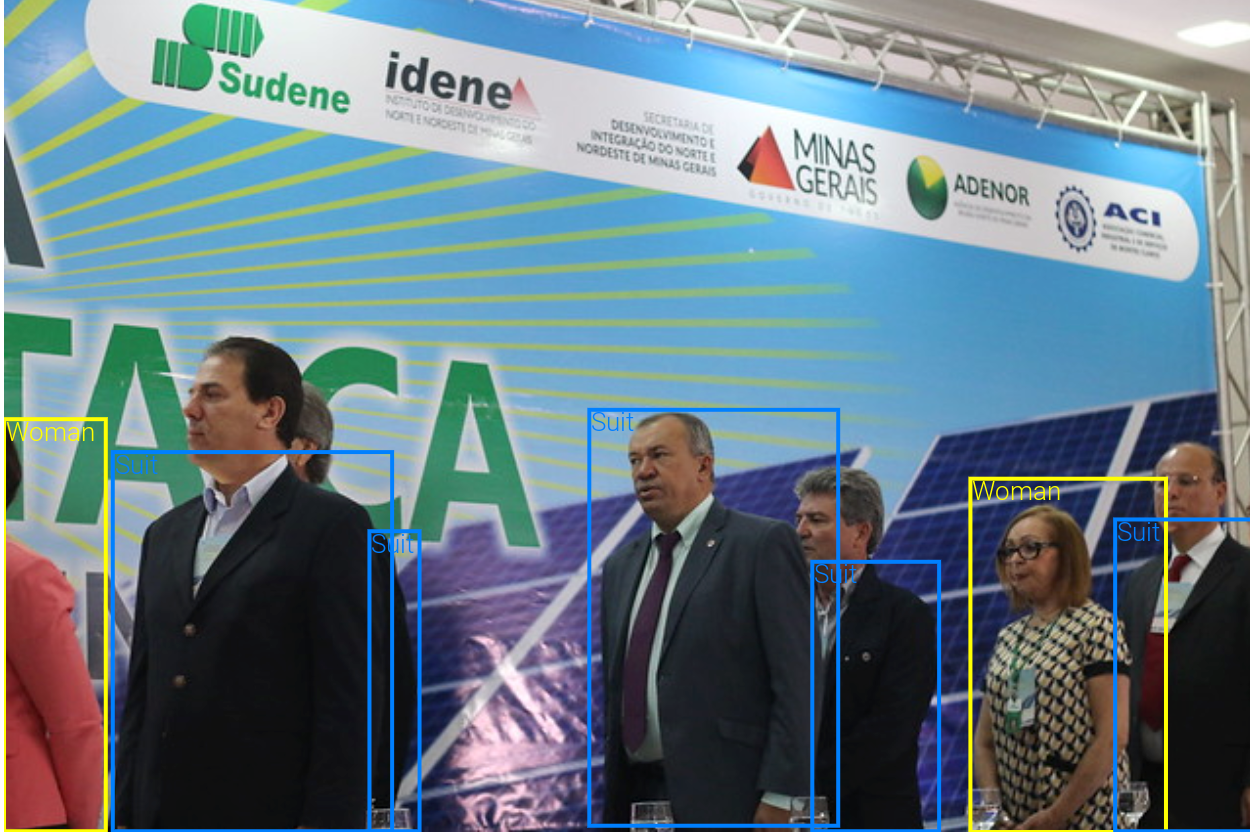}
    \caption{Image with true image-level labels \person, \woman, \emph{suit}.}
    \label{fig:woman_no_man}
\end{figure}

\begin{figure}
    \centering
    \includegraphics[width=0.6\columnwidth]{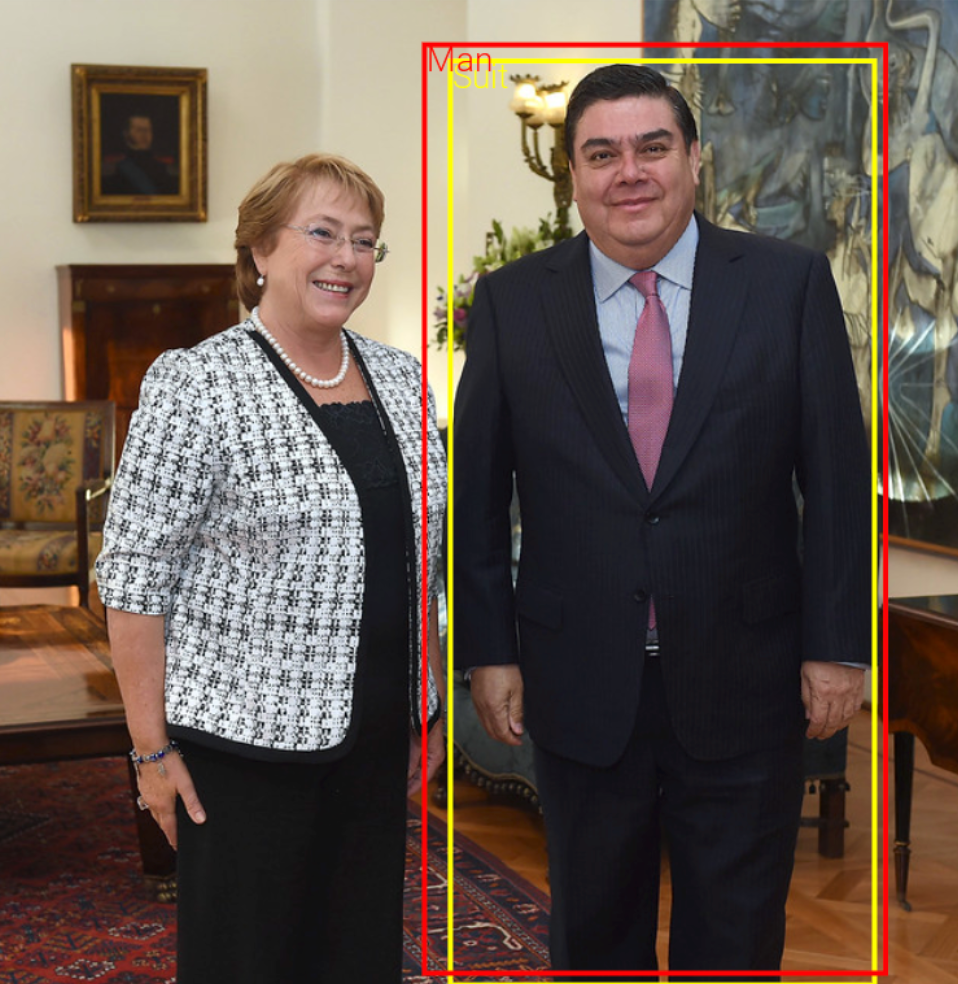}
    \caption{Image with true image-level labels \person, \man, \emph{suit}, \emph{clothing}. 
    }
    \label{fig:man_no_woman}
\end{figure}

\subsubsection{Gender Presentation}
After-the-fact label manipulation can cause individuals of different genders to be treated differently in the resulting annotations.

Consider the case of the image in Figure~\ref{fig:woman_no_man}. The following image-level labels were proposed and marked positive: \person, \woman, and \emph{suit}. Using the hierarchical labeling strategy, annotators were asked to draw boxes around objects of the classes \woman{} and \emph{suit}, but not \person. As a result, the annotations for this image are missing the bounding boxes around the men wearing the suits.

Note that this problem can also happen in the other direction. See Figure~\ref{fig:man_no_woman}, where the image-level label set includes \man{} and \person{}, but not \woman.

The cause of the missing boxes in these examples is obvious: the imbalance of the specificity of the image-level labels proposed for inclusion in the positive set, coupled with the decision to prefer subclasses for box annotations. But subtle omissions can occur even in cases where the positive image-level labels include all five \person-related classes.

Consider another image containing five people: one man, one woman, one girl, one boy, and one person with unidentifiable or nonconforming gender presentation. Assume (unlike the previous example), that the image classifier successfully proposed all relevant labels: \person, \man, \woman, \boy, and \girl. Bounding-box annotators would be instructed to draw boxes around individuals corresponding to the four subclasses. Depending on the specific decisions made by the annotators for this image, this process could potentially miss annotations for people that do not fit into a gender-stereotyped role.

\begin{figure}
    \centering
    \includegraphics[width=0.5\columnwidth]{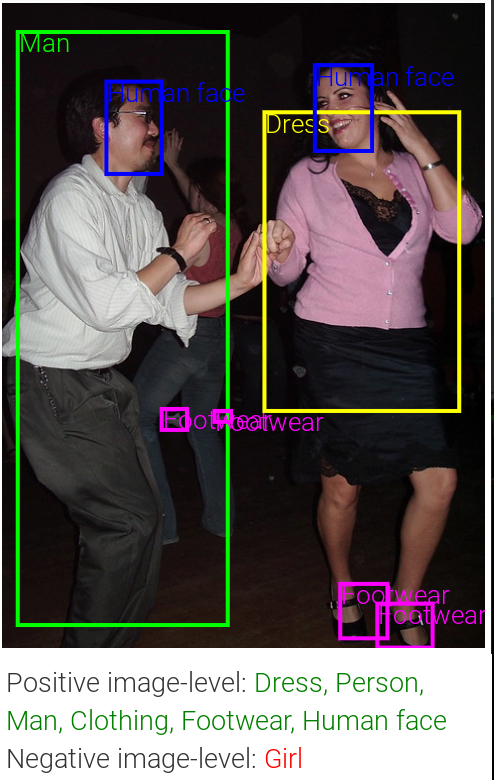}
    \caption{Image with a negative label for \girl{} and no \woman{} replacement}
    \label{fig:negative_label_girl}
\end{figure}

\subsubsection{Age and language}
Similar problems arise in instances where the image classifier and annotators perceive the age of the image subjects differently. In Figure~\ref{fig:negative_label_girl}, we show an example where an age-dependent image-level label (\girl) is labeled as negative with no replacement of the age-appropriate label (\woman). The result is similar to Figure~\ref{fig:man_no_woman}, with a bounding box annotation for only one of the two people in the image.

\begin{figure}
    \centering
    \includegraphics[width=0.8\columnwidth]{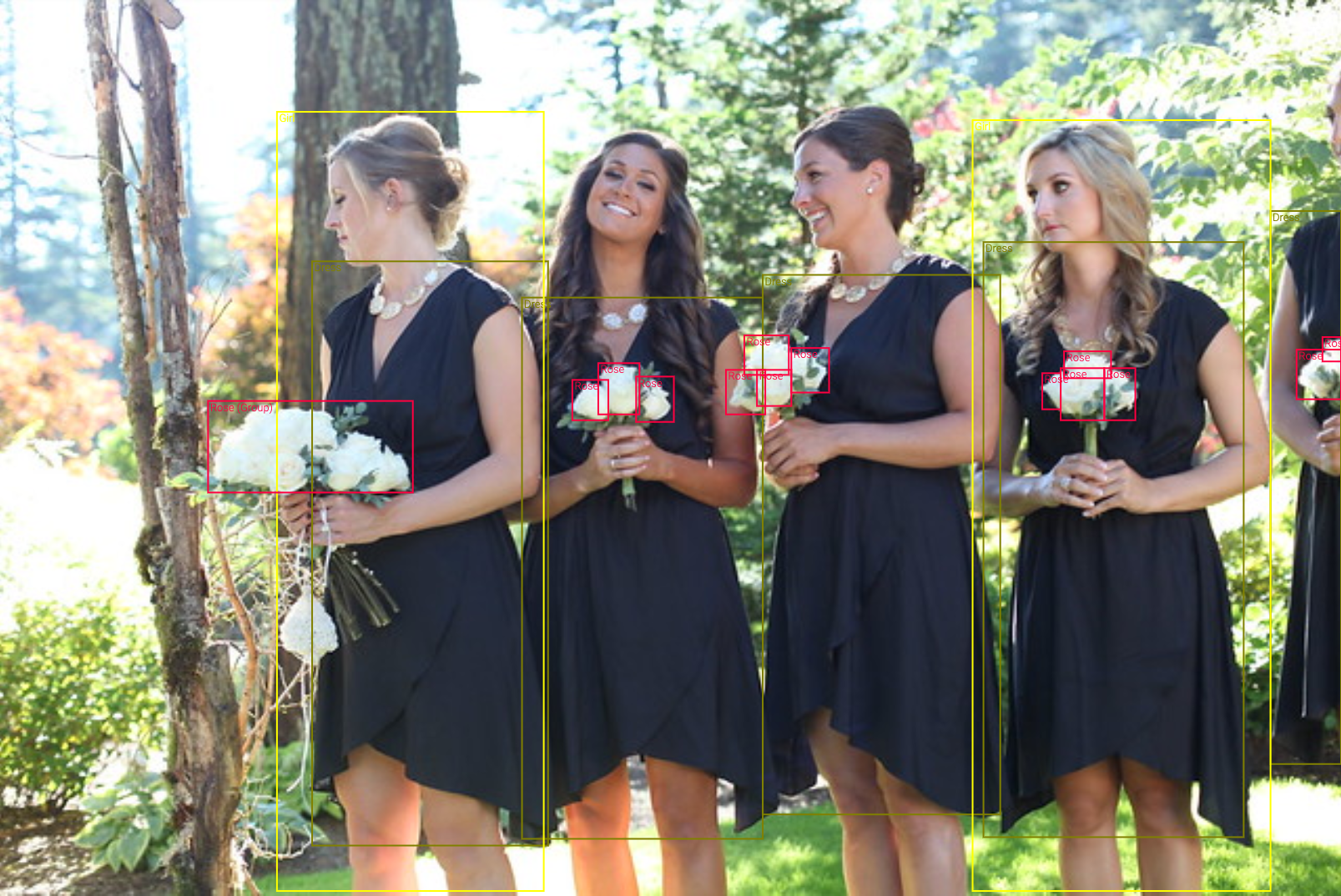}
    \caption{Several people of similar age, only two have bounding boxes drawn around them.}
    \label{fig:one_girl}
\end{figure}

In Figure~\ref{fig:one_girl}, the \girl{} image-level label is considered positive, but an annotator only perceived that label as applying to two people in the image. As a result, several people of similar ages do not have boxes drawn around them.

We observed that phenomena of this type were more likely to cause missed bounding boxes for images with boxes for \girl{} or \woman{} than for images with \man{} or \boy. This may be the result of the difference in how these terms are used and understood by society. The word \girl{} can be used to describe a female of any age while \boy{} typically describes only a child or adolescent.\footnote{For example, Wikipedia defines \girl{} as ``A girl is a young female human, usually a child or an adolescent. When she becomes an adult, she is described as a woman. The term girl may also be used to mean a young woman, and is sometimes used as a synonym for daughter. Girl may also be a term of endearment used by an adult, usually a woman, to designate adult female friends.''~\cite{enwiki:girl} while \boy{} is defined as ``A boy is a young male human. The term is usually used for a child or an adolescent. When a male human reaches adulthood, he is described as a man.''~\cite{enwiki:boy}}

\begin{figure*}[h]
 \centering
 \begin{subfigure}[b]{0.38\textwidth}
     \centering
     \includegraphics[width=\textwidth]{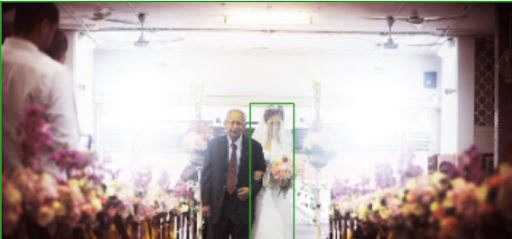}
     \caption{Weddings}
     \label{fig:wedding}
 \end{subfigure}
 \begin{subfigure}[b]{0.3\textwidth}
     \centering
     \includegraphics[width=\textwidth]{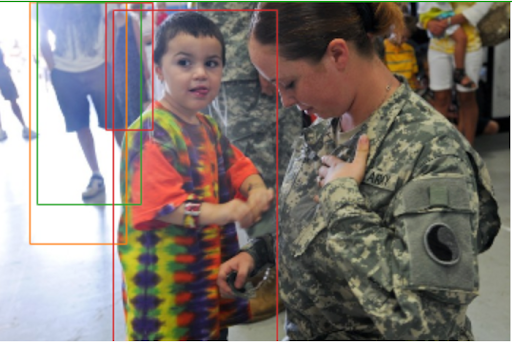}
     \caption{Military}
     \label{fig:military}
 \end{subfigure}
    \caption{Image context and gender norms sometimes dictate which image-level labels are flagged.
    }
    \label{fig:image_context}
\end{figure*}

\subsubsection{Potential correlation with image context}
Because the image content influences which labels are proposed in the classification step of the process (step 2 in Figure~\ref{fig:pipeline}), it affects which boxes are included in the annotations. Partial annotations may be more prevalent, and also potentially more concerning, in scenes that correspond to existing societal stereotypes. Take, for example, the three images in Figure~\ref{fig:image_context}. In traditional heterosexual European and US-style weddings the focus of the day is generally placed on the bride. This emphasis is reinforced in the annotations, with a bounding box for the woman in Figure~\ref{fig:wedding} but not for the man.
As another example, in the US the majority of military personnel are male~\cite{DoD18:Demographics}; a prior that is reinforced in the missing bounding box annotation for the woman in uniform in Figure~\ref{fig:military}.

\subsection{Potential risks of direct use of non-exhaustive annotations}
\label{sec:impact}
A person detector trained on these partial annotations may learn to imitate the non-exhaustive patterns in the training data. As a result, a trained detector may be less likely to correctly detect individuals from certain subpopulations in certain contexts. This is also true for detectors that rely on postprocessing to collapse predictions of \man{}, \woman{}, \boy{}, and \girl{} to \person{}, where correct output would rest on the ability of a specific classifier (e.g., \woman) to generalize to scene contexts that may not have been sufficiently represented in the non-exhaustive training data.

The severity of such errors depends on the use case in question, ranging from annoyances like a less efficient experience searching through personal image collections to highly consequential failures in life safety systems (e.g., pedestrian detection in assistive driving features).

The degree to which trained models are resilient to noisy annotations remains an active area of research. In particular, we do not yet understand the effects that \emph{structured patterns} of the type we observe here have on model performance; typically studies of robustness in the presence of noise make simplifying assumptions and experiment with annotations that are modified or missing uniformly at random~\cite{wu2018soft}. 
Especially concerning here is that the ground truth in the validation and test sets are also non-exhaustive, 
so it may be difficult to detect disparate performance of a model inheriting bias from these annotations. A detailed study of these important questions warrants significant future work that our \emph{MIAP} annotations are designed to support.

\section{More Inclusive Annotations for People}\label{sec:relabeling}

To generate a subset with complete ground truth for the \person{} class and enable related fairness research, we relabel 100K randomly sampled images (70,000 from training and 30,000 from validation/test) that contained at least one of the five \person{} classes (\man, \woman, \boy, \girl, or \person) in the positive image-level set. Annotators were asked to draw boxes around all people in these images. They were then asked to provide two different attributes for each person.

\subsubsection{Perceived Gender Presentation}
Annotators selected either \emph{predominantly feminine}, \emph{predominantly masculine}, or \emph{unknown} to describe the human-perceived gender presentation of an individual based on the visual cues in the image.

We recognize that gender is not binary and that an individual's gender identity may not match their perceived or intended gender presentation. We chose to use \emph{predominantly feminine} and \emph{predominantly masculine} rather than the historical labels \emph{female} and \emph{male} to acknowledge that an individual's gender presentation is complex while maintaining the ability to analyze these labels in the context of the existing image-level and box labels of \emph{man}, \emph{woman}, \emph{boy}, and \emph{girl}. When used in aggregate, these annotations are useful to assess how model performance may differ for people who present gender differently.

We did not include \emph{non-binary} as a class label as it is not possible to label gender identity from images. Gender identity should only be used in situations where participants are able to self-report gender~\cite{GenderGuidelines}.

In an effort to mitigate the effects of unconscious bias on the annotations, we reminded annotators that norms around gender expression vary across cultures and have changed over time. Annotation guidelines included statements like:
\begin{quote}
    No single aspect of a person's appearance ``defines" their gender expression. For example, a person may still present as predominantly masculine while wearing jewelry. Another may present as predominantly feminine while having short hair. Consider the entirety of a person's appearance in the image and avoid rigid definitions based on cultural stereotypes.
\end{quote}
In addition to the written instructions, we provided annotators with visual examples that included people across a variety of ages and from different cultures. These sets included visual examples to counteract common stereotypes (e.g., including images of women with short hair).

Finally, annotators were instructed to select the \emph{unknown} label in situations where there was not enough information in the image to make a determination. Note that annotators could decide to select either \emph{predominantly feminine} or \emph{predominantly masculine} even if an individual's face was not visible.

\subsubsection{Perceived Age Range}
Annotators selected either \emph{young}, \emph{middle}, \emph{older}, or \emph{unknown} to describe the perceived age range of an individual based on their appearance in the image. Annotators were instructed to prefer the older of two categories in situations where there was enough information to form an impression but were unsure of a boundary case. Someone who appears old enough to possibly belong to \emph{middle} should be assigned that attribute label. Someone who appears old enough to possibly belong to \emph{older} should be assigned that attribute label.

Annotators could select \emph{unknown} for perceived age range while selecting one of the other three labels for perceived gender presentation. Similarly they could select \emph{young}, \emph{middle}, or \emph{older} even if they annotated the same bounding box as \emph{unknown} for perceived gender presentation.

\begin{table}
\centering
\begin{tabular}{p{0.25\columnwidth}llll}
    \toprule
    Datasets & Train & Val & Test & Total \\
    \midrule
    images\goodbreak sampled & 70,000 & 7,410 & 22,590 & 100,000 \\
    images with missing boxes & 23,241 & 1,836 & 5,397 & 30,474 \\
    \bottomrule
\end{tabular}
\caption{Distribution of the images selected for relabeling. The second row counts the number of images where the \emph{MIAP} annotations add boxes that do not exist in the original annotations.} 
\label{tab:image_counts}
\end{table}

\begin{table*}
    \centering
    \begin{tabular}{p{0.17\textwidth}llllllll}
        \toprule
         \multirow{2}{*}{Datasets} & \multicolumn{2}{c}{Train} & \multicolumn{2}{c}{Validation} & \multicolumn{2}{c}{Test} &
         \multicolumn{2}{c}{Total} \\
         \cmidrule{2-3} \cmidrule{4-5} \cmidrule{6-7} \cmidrule{8-9} \\
         {} & original & \emph{MIAP} & original & \emph{MIAP} & original & \emph{MIAP} & original & \emph{MIAP} \\
         \midrule
         boxes & 294,695 & 363,165 & 15,398 & 22,060 & 47,777 & 69,106 & 357,870 & 454,331 \\
         \midrule
         \emph{predominantly\goodbreak feminine} & 67,979 & 83,004 & 2,039 & 4,453 & 6,265 & 13,215 & 76,283 & 100,672 \\
         \emph{predominantly\goodbreak masculine} & 127,468 & 141,300 & 3,779 & 7,898 & 12,073 & 24,849 & 143,320 & 174,047 \\
         \emph{unknown}\goodbreak gender presentation & 99,248 & 138,861 & 9,580 & 9,709 & 29,439 & 31,042 & 138,267 & 179,612 \\
         \midrule
         \emph{young} & 16,634 & 22,621 & 1,205 & 1,508 & 3,709 & 4,677 & 21,548 & 28,806 \\
         \emph{middle} & 178,813 & 190,179 & 4,613 & 10,591 & 14,629 & 32,904 & 198,055 & 233,674 \\
         \emph{older} & - & 7,176 & - & 471 & - & 1,376 & - & 9,023 \\
         \emph{unknown} age range & 99,248 & 143,189 & 9580 & 9490 & 29,439 & 30,149 & 138,267 & 182,828 \\
         \bottomrule
    \end{tabular}
    \caption{Comparison counts of boxes across the 100,000 samples from the original Open Images dataset and the \emph{MIAP} dataset. Counts are split between the training set, the validation set, and the testing set, with the totals appearing in the last two columns. Gender presentation and age range presentation for the original set was assumed via the class label (for example, the class \woman{} would have gender presentation \emph{predominantly feminine} and age range presentation \emph{middle}). There was no class for \emph{older} in the original Open Images dataset.}
    \label{tab:raw_counts}
\end{table*}

\section{Analysis}
\label{sec:analysis}

It is important to note the differences between the original Open Images dataset and the new \emph{MIAP} dataset to understand what has improved and what still needs work.

\paragraph{Raw counts}
In Table~\ref{tab:image_counts}, we see that approximately %
30\% of the 100,000 images had additional boxes added during the relabeling effort. This is quite a large percentage of the images, underscoring how often the phenomena described in Section~\ref{sec:societal_norms} did, in fact, occur.

Table~\ref{tab:raw_counts} compares the total number of  boxes in the original annotations to the \emph{MIAP} set, both in aggregate and broken down by attribute label. This analysis assigns a bounding box in the original annotation the two attribute labels according to the class labels (e.g., \woman{} maps to attribute labels \emph{predominantly feminine} and \emph{middle}).

There is a large increase in \emph{unknown} labels (for both gender presentation and age range) in the training set but only a small increase in the validation and test sets. This is explained by the difference in annotation procedure between training and validation/test in the original protocol. The training set used the process described in Section~\ref{sec:pipeline}. For the validation and test sets, annotators drew boxes for all positive image-level labels. %
This means that if the label \person{} appeared in the positive image-level labels for a validation or testing image, the annotator would be asked to draw generic \person{} boxes, regardless of the presence of subclass labels. Boxes for which annotators were not able to assign attribute labels are likely to have been drawn in response to a prompt for \person{} boxes but not drawn when only a more specific class was requested.

We see a significant increase for both \emph{predominantly feminine} and \emph{predominantly masculine} labels across the training, validation, and test splits. 
Although this labeling effort has decreased the gap between \emph{predominantly masculine} and \emph{predominantly feminine} slightly, the dataset is still unbalanced with the majority of gender presentation labels, after \emph{unknown}, belonging to \emph{predominantly masculine}. \emph{Middle} is by far the most common of the age range attribute labels.

\paragraph{Statistical analysis}

In order to measure the association of attribute labels 
with missing boxes we use normalized pointwise mutual information (\emph{nPMI})~\cite{bouma2009normalized}. Pointwise mutual information is defined as
\begin{equation*}
    \textit{PMI}(x,y) = \ln\frac{p(x,y)}{p(x)p(y)}
\end{equation*}
where $x$ and $y$ are discrete random variables. Normalized pointwise mutual information is therefore defined as
\begin{equation*}
    \textit{nPMI}(x,y) = \frac{\textit{PMI}(x,y)}{-\ln p(x,y)}.
\end{equation*}

We use nPMI to calculate the co-occurrences of images with missing boxes and a given label (for example the co-occurrence of images with missing boxes and gender presentation \emph{predominantly feminine}), denoted as $\textit{nPMI}(\text{Missing},\text{label})$, as well as the co-occurrence of images without missing boxes and a given label (for example the co-occurrence of images without missing boxes and gender presentation \emph{predominantly feminine}), denoted as $\textit{nPMI}(\text{Not Missing}, \text{label})$. Using these two co-occurrences we can take the difference $$\textit{nPMI}(\text{Missing},\text{label})-\textit{nPMI}(\text{Not Missing}, \text{label})$$ to find labels that are more likely to co-occur with images that are missing people boxes than images that are not missing people boxes in the original annotations.

Table~\ref{tab:npmi} shows these results in order of significance. Unsurprisingly \emph{unknown} labels for both age range and gender presentation are much more likely to be found in images with missing boxes. This is mostly likely due to the training set labeling process only annotating the most specific \person{} classes. Additionally, the label \emph{predominantly feminine} is much more likely to co-occur with missing boxes than the label \emph{predominantly masculine}, possibly due to the syntactic confusion between \woman{} and \girl{}. The age range labels exhibit the least amount of skew between images with missing boxes and those without. %

\begin{table}[]
    \centering
    \begin{tabular}{ll}
        \toprule
        Label & nPMI difference \\
        \midrule
        gender presentation \emph{unknown} & 0.458 \\
        age range \emph{unknown} & 0.453 \\
        \emph{predominantly feminine} & 0.365 \\
        \emph{predominantly masculine} & 0.251 \\
        \emph{young} & 0.236 \\
        \emph{middle} & 0.156 \\
        \emph{older} & 0.155\\
        \bottomrule
    \end{tabular}
    \caption{nPMI differences between images with missing boxes and images without missing boxes ($\textit{nPMI}(\text{Missing},\text{label})-\textit{nPMI}(\text{Not Missing}, \text{label})$) in order of significance. A high nPMI difference indicates that images with missing boxes were more likely to have new boxes with the given labels.}
    \label{tab:npmi}
\end{table}

\begin{figure*}
\centering
    \begin{subfigure}[b]{0.20\textwidth}
    \centering
    \includegraphics[width=\textwidth]{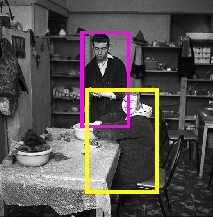}
    \caption{}
    \label{fig:new_1}
    \end{subfigure}
    \begin{subfigure}[b]{0.20\textwidth}
    \centering
    \includegraphics[width=0.8\textwidth]{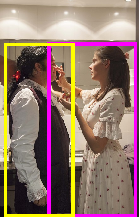}
    \caption{}
    \label{fig:new_2}
    \end{subfigure}
    \begin{subfigure}[b]{0.25\textwidth}
    \centering
    \includegraphics[width=\textwidth]{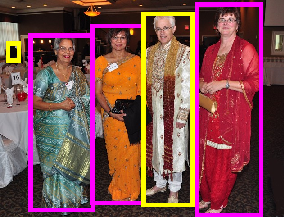}
    \caption{}
    \label{fig:new_3}
    \end{subfigure}
    \begin{subfigure}[b]{0.25\textwidth}
    \centering
    \includegraphics[width=\textwidth]{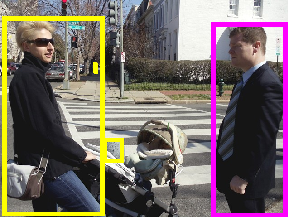}
    \caption{}
    \label{fig:new_4}
    \end{subfigure}
\caption{Examples of new boxes in \emph{MIAP}. In each subfigure the magenta boxes are from the original Open Images dataset, while the yellow boxes are new boxes from the \emph{MIAP} dataset.
}
\label{fig:new_labels}
\end{figure*}

\paragraph{Empirical visualization}
In addition to exploring raw counts and statistical analysis, we visualized the differences between the original Open Images and the \emph{MIAP} annotations. Figure~\ref{fig:new_labels} shows four examples. In each image, the original Open Images had at least one person not annotated. The new \emph{MIAP} dataset has bounding box annotations for all people.

\section{Discussion}
Exhaustively annotated, large-scale visual data is particularly valuable, since it often enables new types of analyses that are ineffectively addressable by partial annotations. With the exhaustive box annotations for the \person{} class, we hope to close this gap for fairness-related analysis of people-centric ML models. We appreciate and encourage its use for tasks such as disaggregating/stratifying metrics along age range and gender presentation slices, in the hope that such analyses will eventually lead to less biased models, classifiers, and workflows that work equitably for all. 

It is important to note that the gender presentation and age range presentation labels, while useful for evaluation purposes, do not represent gender identity or actual age of the people. This data should not be used to create gender and/or age classifiers.
As a further precaution, a gender presentation of \emph{unknown} has been set for any box labeled as \emph{young} in the released dataset\footnote{Analysis included in this paper includes gender presentation annotations for people marked as \emph{young}.}. 

\subsection{Limitations}
The continually evolving and expanding nature of Internet imagery implies that the distribution of Flickr images included within the original Open Images dataset (and consequently, the subset included in this relabeling effort) may not accurately represent the uses of models trained or evaluated using this data. Additionally, recent works have demonstrated that the geographical sampling of Flickr images as well as the use of English as the primary language for dataset construction and taxonomy definition result in inherent cultural bias within the datasets \cite{Vries_2019_CVPR_Workshops}.

While this effort makes initial strides towards relabeling the dataset, there is much work yet to be done. Several parts of the original annotation pipeline for the Open Images dataset were not addressed - including image selection, object class hierarchy, determination of classes that are boxed, and annotation of relationship triplets. Our image selection task rejected images which did not contain at least one \person-subclass label, which may have propagated/amplified any biases that may have existed in the labeling workflow for the original dataset. While this effort has slightly aided in balancing the bounding box annotations, there remains an imbalance towards a gender presentation of \emph{predominantly masculine}.
This relabeling effort did not improve age range imbalance.
Owing to the large amount of manual effort required to exhaustively label each image, this task was also restricted only to certain \person{} labels (\man/\woman/\girl/\boy/\person), but not all \person-related labels that exist in the Open Images taxonomy (e.g. parts of the human body, activities, clothing, etc). Other, non-\person-related classes in the Open Images taxonomy may also benefit from similar analysis and exhaustive relabeling, but were not addressed in this paper. Relationship triplets were not re-annotated, so they only exist for the subset of boxes that were included in the original dataset.

We acknowledge that there are trade-offs between providing the capability to do fairness analysis based on societal concepts of perceived gender, and perpetuating binary definitions of gender. Future work could include analysis of objective attributes that can confound computer vision models, such as makeup or facial hair.

\section{Conclusion}
Dataset construction, annotation procedures, fairness analysis and techniques for modeling under noisy ground truth conditions are all continually evolving areas of research.

In this paper, we have presented a new annotation procedure and the resulting annotations for a subset of the Open Images dataset, titled \emph{MIAP (More Inclusive Annotations for People)}. The new annotation procedure results in more annotations for the \person{} class, as well as annotations for gender presentation and age range presentation. The annotation procedure was additionally designed to enable fairness analysis.

Our analyses of the two annotation pipelines and resulting annotation patterns demonstrate the importance of understanding annotation design decisions and their consequences. We have discussed how annotation procedures can be seen to interact with societal norms, language, and image context, as well as other data patterns, here resulting in differences for the \person{} class and subclass labels. The original annotations and new annotations for Open Images are both useful in different situations since they vary in scope and coverage, or they can be used as complementary information. We hope that the findings in this paper can inform usage of the current dataset as well as future dataset annotation efforts.

We encourage researchers to use our new \emph{MIAP} annotations to incorporate fairness analysis into their research. Given the exhaustive nature of our updated annotation procedure, researchers can expect more robust fairness evaluations.

\section*{Image citations}
All images used in this paper are licensed under Creative Commons By 2.0 from Open Images. References for each image are as follows:
\begin{itemize}
    \item Figure~\ref{fig:woman_no_man}: thalisantunes, \url{https://www.flickr.com//photos//thalisantunes//21322592792//}
    \item Figure~\ref{fig:man_no_woman}: Gobierno de Chile, \url{https://www.flickr.com/photos/gobiernodechile/16822006848/}
    \item Figure~\ref{fig:negative_label_girl}: Waifer X, \url{https://www.flickr.com/photos/waiferx/85365663/}
    \item Figure~\ref{fig:one_girl}: Alex Erde, \url{https://www.flickr.com/photos/alexerde/9658256603/}
    \item Figure~\ref{fig:wedding}: sapex, \url{https://www.flickr.com/photos/sapexoxo/4884905970/}
    \item Figure~\ref{fig:military}: 29th Combat Aviation Brigade Public Affairs,  \url{https://www.flickr.com/photos/29thcab/6155989139/}
    \item Figure~\ref{fig:new_1}: Grigory Kravchenko, \url{https://www.flickr.com/photos/grisha_21/3464190158/}
    \item Figure~\ref{fig:new_2}: Quincena Musical, \url{https://www.flickr.com/photos/quincenamusical/15026832341}
    \item Figure~\ref{fig:new_3}: Simon Fraser University - Communications \& Marketing, \url{https://www.flickr.com/photos/sfupamr/6238487408/}
    \item Figure~\ref{fig:new_4}: Emma Racila, \url{https://www.flickr.com/photos/94605075@N05/8617525388/}
\end{itemize}

\bibliography{references} 
\bibliographystyle{ACM-Reference-Format}

\end{document}